%% file: main.tex
\documentclass[10pt,twocolumn,letterpaper]{article}

\usepackage{iccv}              

\definecolor{iccvblue}{rgb}{0.21,0.49,0.74}
\usepackage[pagebackref,breaklinks,colorlinks,allcolors=iccvblue]{hyperref}
\usepackage{graphicx}

\usepackage{amsmath}
\usepackage{amssymb}
\usepackage{colortbl}
\usepackage{graphicx}
\usepackage{xcolor}
\usepackage{multirow}
\usepackage{booktabs} 
\usepackage{amssymb}
\usepackage{pifont}
\usepackage{algorithm}
\usepackage{algpseudocode}
\usepackage{algorithmicx}
\usepackage[dvipsnames]{xcolor}
\usepackage{makecell}
\usepackage{rotating}

\usepackage{arydshln}

\title{Diversity Covariance-Aware Prompt Learning for Vision-Language Models}

\author{{Songlin Dong\textsuperscript{\rm 1}} \quad
Zhengdong zhou\textsuperscript{\rm 1} \quad 
Chenhao Ding\textsuperscript{\rm 2} \quad  
Yuhang He\textsuperscript{\rm 2} \\  
Xinyuan Gao\textsuperscript{\rm 1} \quad 
Alex Kot \textsuperscript{\rm 3} \quad 
Yihong Gong\textsuperscript{\rm 1,} \textsuperscript{\rm 2}\quad 
\vspace{0.2em} \\
\textsuperscript{\rm 1}School of Software Engineering, Xi’an Jiaotong University \\
\textsuperscript{\rm 2}College of Artificial Intelligence, Xi’an Jiaotong University \\
\textsuperscript{\rm 3}Nanyang Technological University \\
}

\begin{document}
\maketitle

\begin{abstract}
Prompt tuning can further enhance the performance of visual-language models across various downstream tasks~(\textit{e.g.}, few-shot learning), enabling them to better adapt to specific applications and needs. In this paper, we present a \textbf{D}iversity \textbf{C}ovariance-\textbf{A}ware framework that learns distributional information from the data to enhance the few-shot ability of the prompt model. First, we propose a covariance-aware method that models the covariance relationships between visual features and uses anisotropic Mahalanobis distance, instead of the suboptimal cosine distance, to measure the similarity between two modalities. 
We rigorously derive and prove the validity of this modeling process. 
Then, we propose the diversity-aware method, which learns multiple diverse soft prompts to capture different attributes of categories and aligns them independently with visual modalities. This method achieves multi-centered covariance modeling, leading to more diverse decision boundaries. Extensive experiments on 11 datasets in various tasks demonstrate the effectiveness of our method. 
\end{abstract}

\vspace{-0.4cm}
\section{Introduction}
Recently, vision-language models (VLMs), such as CLIP \cite{clip}, have achieved significant success in various applications within computer vision, including image recognition~\cite{coop,gao2024clip}, action understanding~\cite{tevet2022motionclip}. However, despite VLMs demonstrating promising generalization capabilities and transferability across various downstream tasks, adapting the knowledge gained during pre-training to specific downstream tasks~(\textit{e.g.}, few-shot learning) remains a significant research challenge, as these models are typically massive sizes, and retraining them consumes substantial computational resources.

\begin{figure*}[htb!]

\begin{center}
    \includegraphics[width=0.93\linewidth]{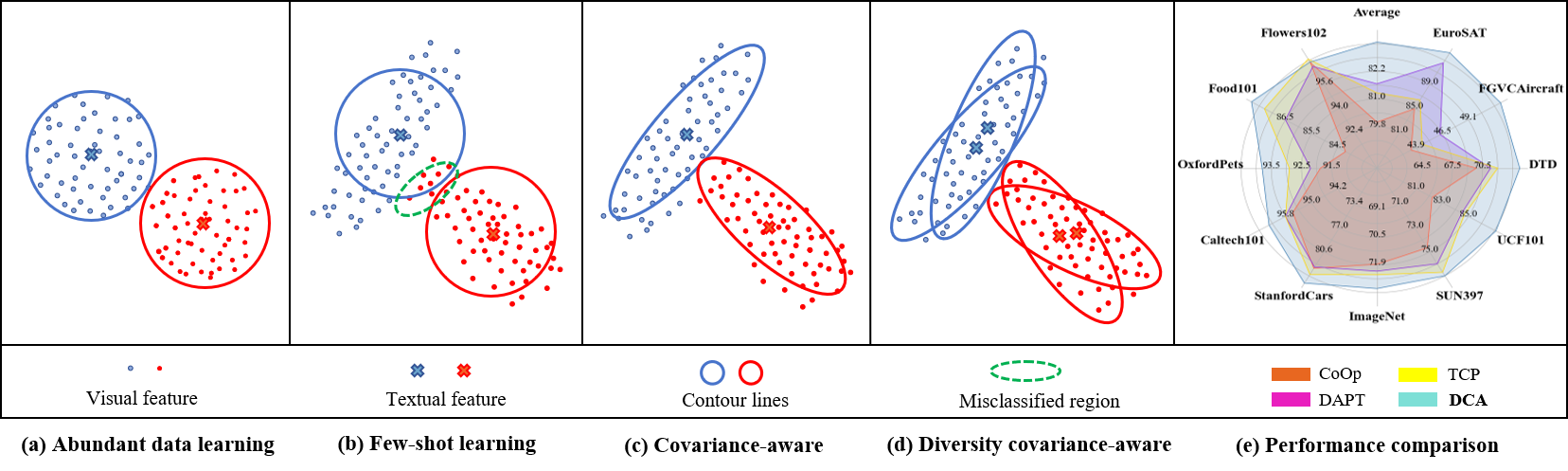} 
\end{center}
\setlength{\belowcaptionskip}{-0.2cm} 
\setlength{\abovecaptionskip}{-0.3cm} 
\caption{Illustration of feature space distribution and performance comparison. (a) When training data is abundant, DNN learns a good isotropic spherical feature space~\cite{ncm1}, and thus, isotropic metrics (such as cosine or Euclidean distance) can be effectively applied.
(b) However, in few-shot tasks, isotropic feature space becomes challenging~\cite{kumar2022gdc,goswami2024fecam}, and isotropic metrics can lead to misclassified regions (as shown in the green area).
(c) We propose the CA method, which extracts the distribution information of the data through covariance modeling and uses anisotropic Mahalanobis distance to measure the distance. (d) The DA method is designed to combat overfitting and capture different attributes of categories, maintaining more diverse decision boundaries. (e) The DCA method surpasses state-of-the-art methods on 11 diverse datasets.}
\label{fig:head}
\end{figure*}

The mainstream approaches for improving the transfer ability of VLMs apply the \textit{prompting} technique.
Earlier methods \cite{prmopt_mu1,radford2019language} rely on manually crafting templates based on prior human knowledge (\textit{e.g.}, ``a photo of a [label]"). However, designing suitable manual prompts requires domain expertise and is a labor-intensive engineering technique. Thus, recent approaches have proposed \textit{soft prompts} for prompt tuning, which involves optimizing \textit{learnable vectors} in either the text or visual modality. Out of these, CoOp~\cite{coop} and VPT~\cite{VPT} fine-tune CLIP by optimizing a continuous set of prompt vectors within its language branch or image branch, respectively. Since tuning prompts in only one branch of CLIP limits the flexibility to adjust visual and textual representation spaces in downstream tasks~\cite{maple}, subsequent approaches~\cite{maple,DAPT,tcp2,lamm} began to optimize both branches. For example, MaPLe~\cite{maple} proposes multi-modal prompts for both vision and language branches to enhance alignment. DAPT~\cite{DAPT} optimizes both modalities' prompts through metric learning to adjust the arrangement of the latent space. 

Despite their decent performance, existing methods~~\cite{maple,DAPT,tcp2,lamm} have a significant drawback due to the nature of few-shot learning, namely, neglecting the impact of the feature space distribution. When deep neural networks learn from limited data, they are unable to obtain the uniform and isotropic spherical representation~\cite{ncm1} that can be achieved with abundant data, which feature distribution tends to be heterogeneous~\cite{kumar2022gdc} (as shown in Fig.~\ref{fig:head} (a) and (b)). Consequently, during the testing phase, existing methods that rely on isotropic distances (such as cosine or Euclidean distance) to measure the distance between text and image features are suboptimal. From Fig.~\ref{fig:head} (b), we can observe the misclassification region within the green curve. This occurs because the impact of the true feature space distribution is overlooked, which weakens the discriminative ability between features of different categories.

In this paper, we propose the \textbf{d}iversity \textbf{c}ovariance-\textbf{a}ware~(DCA) framework as a way to learn distribution information from the data, which can effectively adapt the pre-trained VLM to downstream few-shot tasks. First, \textit{covariance-aware}~(CA), we demonstrate that a Bayesian classifier~\cite{kumar2022gdc,goswami2024fecam} can be used by modeling the covariance relationships among visual features and utilizing text features as the average vector of these visual features. Specifically, we calculate the covariance matrix for each class from the corresponding visual feature of the training samples and replace the suboptimal cosine distance with the anisotropic Mahalanobis distance when measuring the distance between the two modalities. By considering covariance when calculating distances, the prompt model can better capture more complex class structures in high-dimensional feature spaces, thereby enhancing discrimination between features of different classes, as illustrated in Fig~\ref{fig:head} (c). Furthermore, we enhance the commonly used loss function~\cite{xie2022zero, DAPT} by integrating covariance modeling, thereby improving the ability to adjust intra-class relationships. Second, \textit{diversity-aware}~(DA), we learn multiple distinct and informative soft prompts within the text encoder to capture the diverse attributes of categories and align them with the distribution of the visual modality. Concretely, we extend multiple text prompts of varying lengths and model them \textit{independently} as mean vectors in the covariance modeling. This multi-center covariance modeling effectively alleviates overfitting to limited samples, thereby better simulating the true feature space distribution~(See in Fig.~\ref{fig:head} (d)). Also, we introduce the text separation loss to ensure that different text prompts for the same class capture more comprehensive information.

To verify the effectiveness of our method, we perform experiments across different datasets, focusing on various tasks and we achieve significant performance improvements over the SOTA approaches, shown in Fig.~\ref{fig:head} (e). The main contributions of this paper are summarized as follows:

\begin{itemize}
    \item We propose the diversity covariance-aware (DCA) framework, which learns distributional information from the data to enhance the few-shot ability of the prompt model.
    \item We suggest the covariance-aware~(CA) method that uses anisotropic Mahalanobis distance instead of cosine distance to measure two modalities by modeling the covariance relationships between visual features.
    \item We propose a diversity-aware (DA) method that learns multiple independent soft prompts within the text encoder to achieve multi-centered covariance modeling, leading to more diverse decision boundaries. 
    \item Extensive comparative results and ablation studies across various datasets demonstrate the effectiveness of the proposed DCA framework. 
\end{itemize}


\section{Relate work}
\noindent\textbf{Vision-Language Models.}
Recently, using extensive image-text data in pre-trained vision-language models (VLMs) to explore the semantic correspondence between vision and language has become a trend~\cite{clip,align,wang2022vlmixer}. Among these, CLIP~\cite{clip} aggregates 400 million image-text pairs and employs a contrastive objective to learn vision-language representations. ALIGN~\cite{align} leverages a dataset of 1.8 billion noisy image-text pairs. Then, several approaches~\cite{clip-wang,goel2022cyclip} have emerged that enhance the capabilities of VLM by refining the latent space. These VLMs demonstrate impressive performance across various downstream tasks~\cite{coop,gao2024clip,tevet2022motionclip,VQA1,VQA3}. Currently, the mainstream solutions for visual cognition tasks are \textit{prompt tuning}~\cite{coop,DAPT} and \textit{adapter tuning}~\cite{tip-ada,gao2024clip,vlm-adapter2,clap,tip-mul}. This paper focuses on the former, \textit{i.e.} prompt learning, aiming for fewer parameters and greater efficiency.

\noindent\textbf{Prompt Learning.}
Prompt learning reformulates downstream tasks as pre-training tasks using prompts, effectively adapting pre-trained knowledge by minimizing domain shifts. Early approaches~\cite{prmopt_mu1,radford2019language} relied on manually crafting templates based on prior human knowledge. Additionally, methods using mining techniques and gradient-based approaches~\cite{jiang2020can,shin2020autoprompt} have been developed to discover suitable prompt templates automatically. In addition to hard prompt methods, computer vision has actively explored soft prompts, which involve optimizing learnable vectors in either the text or visual modality. Out of those, CoOp~\cite{coop} fine-tunes CLIP by optimizing a continuous set of prompt vectors within its language branch. CoCoOp~\cite{Cocoop} learns a lightweight network to generate input condition vectors for each image. Other approaches, such as ProDA~\cite{ProDA}, PLOT~\cite{PLOT}, and KAPT~\cite{KRPT} employ multiple text prompts or introduce external information, such as Wikipedia or GPT-3~\cite{GPT3}, to enrich the descriptions of text prompts. Recently, some methods have proposed real-time adjustments to the visual modality. For instance, VPT~\cite{VPT} enhances visual ViT with learnable vectors for prompt tuning. MaPLe~\cite{maple} introduces a multi-modal prompt learning approach by jointly learning hierarchical prompts in both. DAPT~\cite{DAPT} adjusts the arrangement of the latent space to align the two modalities by multiple loss functions. LAMM~\cite{lamm} proposes a hierarchical loss that addresses the discrepancy in class label representations between VLMs and downstream tasks. Also, TCP~\cite{tcp2} proposes text-based class-aware prompt tuning, explicitly incorporating prior knowledge about classes to enhance their discriminability.

\begin{figure*}[htb!]
\setlength{\belowcaptionskip}{-0.8cm}
\begin{center}
\centerline{\includegraphics[width=0.92\linewidth]{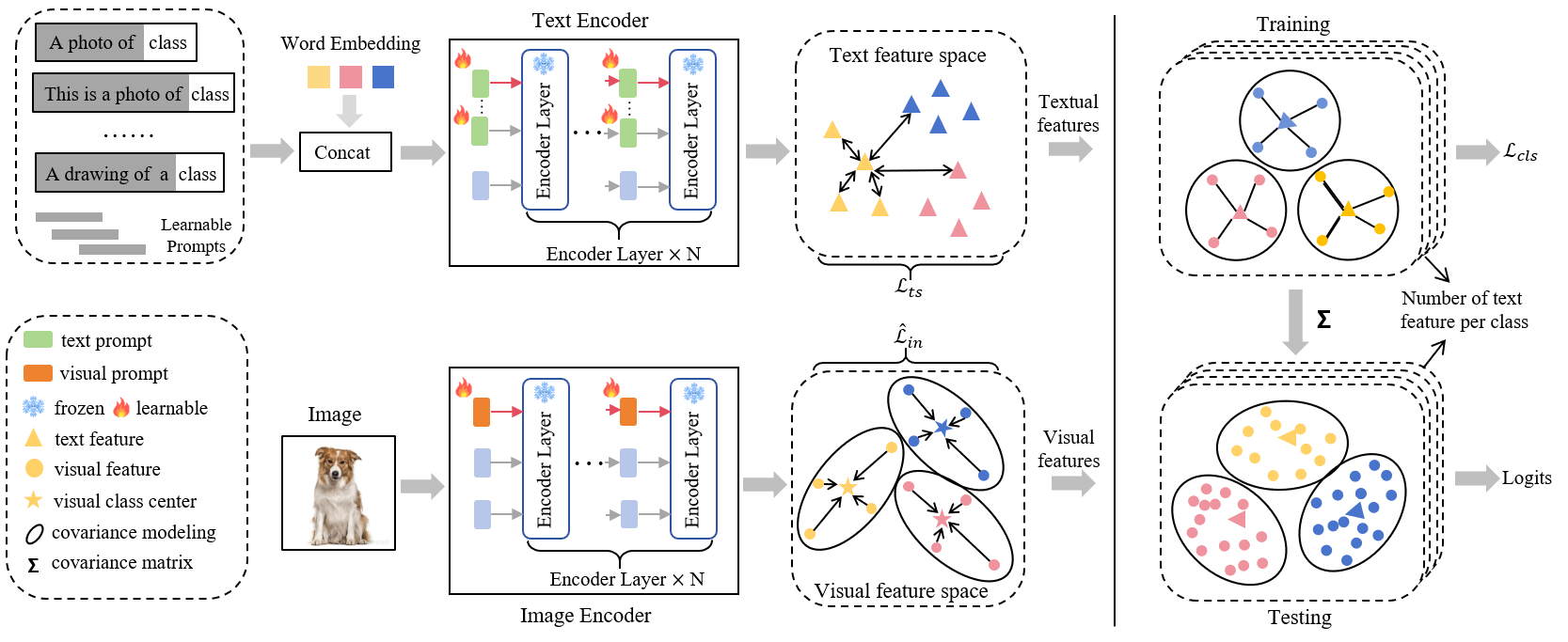}}
\setlength{\abovecaptionskip}{-0.05cm} 
\caption{Overview of the architecture of DCA. DCA first uses multiple prompts to describe each class, generating multiple sets of text features through a text encoder. The images are also encoded into a set of visual features. During training, we independently compute the $\mathcal{L}_{cls}$ between multiple sets of text features and visual features. Meanwhile, we model the covariance relationships between visual features and treat the text features as their average vector, enabling the use of Mahalanobis distance to measure the similarity between text and visual feature modalities during testing. Moreover, we introduce $\mathcal{L}_{ts}$ and improve $\hat{\mathcal{L}}_{in}$ to optimize text and visual feature spaces. The detailed training and testing algorithms are provided in the \textbf{appendix}.}
\label{fig:overview}
\end{center}
\end{figure*}

\section{Method}

\subsection{Preliminary: Prompting for Baseline}
The CLIP~\cite{clip} model is a typical dual-encoder architecture consisting of an \textit{image encoder} $z(\cdot)$ and \textit{text encoder} $g(\cdot)$, which transforms image and text labels into visual and text features, respectively. The training objective is to align the image-text feature pair with contrastive learning on a large-scale dataset.

\noindent\textbf{Text modality.} Soft prompt tuning has been widely applied in the CLIP text encoder. Specifically, the text prompt \( p \) consists of learnable vectors \( \mathbf{w} = \{w_l \mid l = 1, \ldots, L\} \) combined with class name. Each vector has the same dimension as the original word embedding, and \( L \) represents the length of the context words. The text prompt can be formulated as $p = [w_1, w_2, \cdots, w_L, \text{class}]$, where $\text{class}$ is the word embedding of the class. With $p$ as the input, the text encoder $g(\cdot)$ outputs the text feature as $ \mu= g(p)$.

\noindent\textbf{Visual modality.} In visual cognition, some methods propose applying soft prompt tuning in the visual encoder, specifically using visual prompt vectors \( \mathbf{v} = \{v_l \mid l = 1, \ldots, L\} \). These vectors are typically embedded between the class token \( \text{CLS} \) and the image tokens $E$. Similar, the visual prompt can be formulate as $q = [\text{CLS}, v_1, v_2, \cdots, v_L, E]$. With $q$ as the input, the image encoder $z(\cdot)$ outputs the visual feature as $f = z(q)$.

\noindent\textbf{Prompt Tuning.} Given the normalized image-text feature pair \((f,u)\), the prediction probability is calculated by using softmax with the cosine similarity between the image feature and the corresponding text feature of the image categories as follows:
\begin{equation}
\label{base}
p(c=y \mid x) = \frac{\exp(\cos(f, u_y)/\tau)}{\sum_{j=1}^{N} \exp(\cos(f, u_j)/\tau)},
\end{equation}
where \( \cos(f, \mu_y) \) denotes the cosine distance, $\tau$ is the temperature, \(N\) is the total number of image classes and $\mu_y$ represents the text feature of the class label $y$. Then, we can optimize the weight of $\mathbf{w}$ and $\mathbf{v}$ with cross-entropy loss between the prediction probability and the labeled target.

\subsection{Covariance-Aware Prompt Learning. }
\noindent\textbf{Modeling Features with Covariance Relations.}
Considering the feature distribution heterogeneity due to data scarcity, we adopt an anisotropic Gaussian distribution \(\mathcal{N}(\mu_y, \Sigma_y)\) to model the feature distribution of class \( y \) under the few-shot setting. The probability of a feature \( x \) belonging to class \( y \) can be expressed as follows:
\begin{align}
\label{md}
P(x\mid c=y) = k\exp\left(-\frac{1}{2} (x - \mu_y)^T \Sigma^{-1}_y (x - \mu_y)\right)
\end{align}
where $k=\frac{1}{\sqrt{(2\pi)^D |\Sigma_y|}}$ is a constant and   $\mu_y=\frac{1}{N_{i}}\sum_{n=1}^{N_{i}}x_i$ is the average feature vector of all feature of class $y$. $x, \mu_y \in R^D$ and $\Sigma^{-1}_y$ is an arbitrary positive definite matrix. Based on the principle of maximization and Bayes' theorem, we can arrive at the following optimal Bayesian classifier:

\begin{equation}
\label{basi}
\arg \max_{y} P(Y \mid X) = \arg \max_{y} \log P(X \mid Y),
\end{equation}
where the detailed intermediate derivation process is provided in the \textbf{appendix}. Subsequently, by substituting Eq.~\ref{md} into the right side of Eq.~\ref{basi} and eliminating the irrelevant constant \(k\), we obtain the following formulation:

\begin{align}
\arg \max_{y} P(Y \mid X) =  \arg \min_{y} \, (x - \mu_y)^T \Sigma^{-1}_y (x - \mu_y),  
\label{argmd}
\end{align}
where \((x - \mu_y)^T \Sigma_y^{-1} (x - \mu_y)\) is precisely the squared Mahalanobis distance. For convenience, \((x - \mu_y)^T \Sigma_y^{-1} (x - \mu_y)\) can be denoted as \(d_m(x, \mu_y)\). 
Based on the above derivation, we convert the Bayesian classification problem into the problem of minimizing the Mahalanobis distance between features and class means. Therefore, we can solve the probability problem using the Mahalanobis distance

\noindent\textbf{Covariance Modeling in Visual-Language Models.}
Extensive literature researches indicate that the key to prompt tuning is achieving feature space alignment between the two modalities through learnable vectors, specifically the alignment between visual features \(f\) and text features \(\mu\). In the few-shot learning task, multiple visual features \(f\) gradually align with a single text feature \(\mu\) \textbf{during training}, and the category of a visual feature \(f\) is determined by its distance to different text features \(\mu\) \textbf{during testing}. Therefore, we can treat the text feature \(\mu\) as the average vector of visual features from the same class to establish covariance relationships. 

Specifically, for any pair of visual-text features \((f, \mu_y)\), we can calculate the Mahalanobis distance between them as:
\begin{align}
      d_m(f, u_y) =   (f-\mu_y)^T\mathbf{\Sigma}_y^{-1}(f-\mu_y).
\label{ma_base}
\end{align}

However, in a few-shot setting, the number of samples for a class is much smaller than the feature dimension, making the covariance matrix \(\Sigma\) non-invertible and impossible to compute. Therefore, we perform the covariance shrinkage~\cite{kumar2022gdc,goswami2024fecam} to obtain a full-rank matrix as follow:
\begin{equation}
    \Sigma_{s} =  \Sigma+  \gamma _1V_mI+\gamma _2V_n(1-I),
\end{equation}
where \(V_n\) is the average diagonal variance and \(V_m\) is the average off-diagonal covariance of \(\Sigma\). Then, to prevent excessive differences in covariance matrices across different classes, we normalize each class full-rank covariance matrix $\Sigma_{s}$ to obtain \(\hat{\Sigma}_{s}\). Thus, Eq~\ref{ma_base} can be improved as:
\begin{align}
      \hat{d}_m(f, u_y) =   (f-\mu_y)^T\mathbf{(\hat{\Sigma}}_y)_s^{-1}(f-\mu_y).
\label{ma}
\end{align}

According to Eq.\ref{argmd}, we treat maximizing the probability as a problem of minimizing the Mahalanobis distance. Therefore, combining Eq.\ref{base}, \ref{argmd}, and \ref{ma}, the prediction probability formula for the VLMs after covariance modeling in the visual and text feature spaces is as follows:
\begin{equation}
\label{mapr}
p(c=y \mid x) = \frac{\exp(\tau / (\hat{d}_m(f, u_y)+\varepsilon))}{\sum_{j=1}^{N} \exp(\tau / (\hat{d}_m(f, u_j)+\varepsilon))},
\end{equation}
where \(\varepsilon\) is a small positive number for numerical stability. Then we optimize the model for image classification on the downstream data $D$ with cross-entropy loss, ${\mathcal{L}}_{cls}$, as:
\begin{equation}
\label{cls}
    {{\mathcal{L}}_{cls}}=-\tfrac{1}{{N_D}}\sum\limits_{i=1}^{{{{N}}_{D}}}{({y_i}\cdot \log ({p_i})}),
\end{equation}
where \({{{N}}_{D}}\) is the image number of dataset ${\mathcal{D}}$, \(y_i\) and \(p_i\) are the true label and the predicted for input image \({{\boldsymbol{x}}_{i}}\). Considering the properties of the Bayesian classifier, we use Mahalanobis distance only during the testing~(only compute the final epoch per-class covariance matrices). Thus, during training, we simplify the covariance matrix in Eq. 7 to the identity matrix (i.e., use Euclidean distance $d_e$). Moreover, in cases with too few training samples~(\textit{e.g.} 1-shot), we treat all categories in the dataset as a single category to calculate a unified covariance matrix to model the feature space.

\noindent\textbf{Covariance Modeling in Intra-class Loss.}  
In classification tasks, some approaches~\cite{tao2020few,wei2019marginal} use loss functions to minimize intra-class feature distances, thereby enhancing the discriminative power of deep features. Such loss functions are also applied in the VLM tasks~\cite{DAPT,xie2022zero} as follows:

\begin{figure*}[t]
\setlength{\abovecaptionskip}{-0.1cm} 
\setlength{\belowcaptionskip}{-0.4cm} 
\begin{center}
    \includegraphics[width=0.92\textwidth]{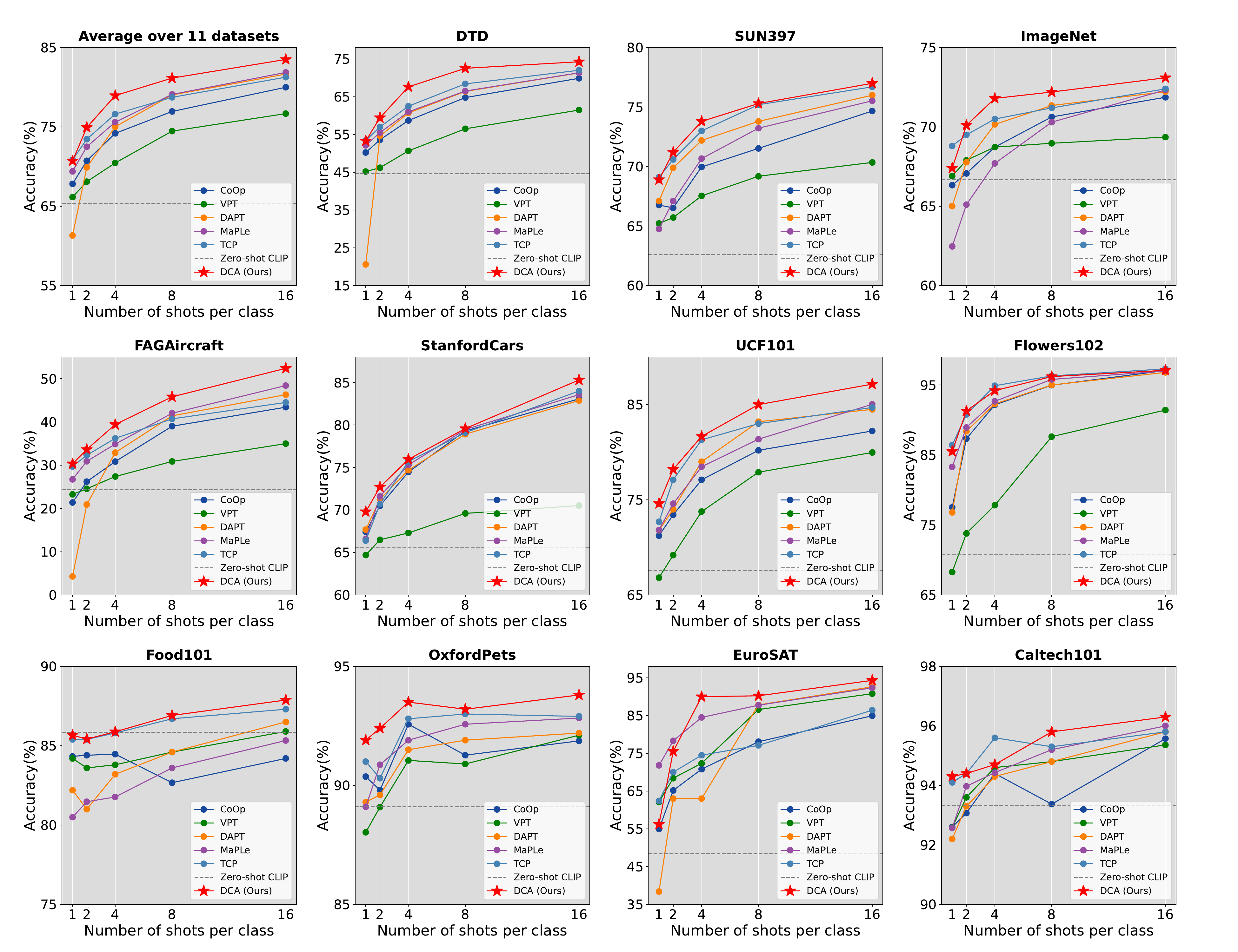}
\end{center}
\caption{Comparison results (\%) of few-shot learning on 11 datasets. We compared our method with several SOTA prompt learning methods and consistently demonstrated significant performance improvements across most datasets. More detailed comparison results and variance of the DCA method are provided in the \textbf{appendix}.}
\label{fig:coco}
\end{figure*}

\begin{equation}
\mathcal{L}_{in} = \sum_{i} \left\| f_i - c_{y_i} \right\|_2^2,
\end{equation}
where $f_i$ is the visual feature, $c_{y_i}$ is the center of features. Based on the assumption of anisotropic Gaussian distributions in the feature space for few-shot tasks, we model the covariance in the intra-class loss $\mathcal{L}_{in}$ as follows:
\begin{equation}
\label{intra}
{\hat{\mathcal{L}}}_{in}=\sum_{i}(f_i-c_{y_i})^{T} \mathbf{(\hat{\Sigma}}_{y_i})_s^{-1} (f_i-c_{y_i}),  
\end{equation}
where $\mathbf{(\hat{\Sigma}}_{y_i})_s^{-1}$ is the covariance matrix of class $y_i$. During training, we compute the covariance matrix only at the first epoch, which significantly reduces the computational burden and effectively maintains the generalization ability of the original CLIP, preventing overfitting.

\subsection{Diversity-Aware Prompt Learning}
\label{diversity}

Even though soft prompts are highly robust, learning a single sentence is insufficient to represent a class. Thus, we learn multiple distinct soft prompts in the text encoder to capture different attributes of a category and its visual representation distribution.

Specifically, compared to the baseline, we extend to multiple learnable vectors \( \mathbf{w}^1, \ldots, \mathbf{w}^M \) for each class and create \( M \) text prompts \( p^1, \ldots, p^M \).
Given an input image \( x \), we obtain the normalized image feature \( f \). We then input \( M \) text prompts into the text encoder to get \( M \) text features \( u^1, \ldots, u^M \), resulting in \( M \) pairs of text features \((f, u^m)\). We then calculate the prediction probabilities for each pair using Eq.~\ref{mapr}, compute the cross-entropy loss for each pair using Eq.~\ref{cls}, and sum these losses to obtain the final loss. Notably, during training, we compute the loss function independently for different text prompts of the same class. This encourages the text prompts to capture various aspects of the visual features, leading to a more comprehensive representation of the class. During testing, we compute the prediction probabilities for the \( M \) text-image pairs using Eq.~\ref{mapr} and average these probabilities to determine the final result.

Moreover, we introduce a text separation loss $\mathcal{L}_{ts}$ to ensure that text prompts for the same class capture more comprehensive information and to address issues with visual feature alignment caused by small distances between text features of different classes. Specifically, we obtain \( M \times C \) text features during training, where \( C \) is the total number of classes. We define these features as the set \( A \). For each pair of \( u_i \) and \( u_j \) randomly selected from the instance set \( A \), The text separation loss can be formulated as:
\begin{equation}\label{eq:out}
\mathcal{L}_{ts} = -\sum_{i\neq j}^{} ||u_i, u_j||^2_{2}.
\end{equation}

\subsection{Object Optimization.}
We optimize the weights of the prompts \( \mathbf{w} \) and \( \mathbf{v} \) using three loss functions: (1) the classification loss \( {\mathcal{L}}_{cls} \) in Eq.~\ref{cls}, (2) the Mahalanobis intra-class loss \( \hat{\mathcal{L}}_{in} \) in Eq.~\ref{intra}, and (3) the text separation loss \( \mathcal{L}_{ts} \) in Eq.~\ref{eq:out}. The total loss function is then obtained as follows:

\begin{equation}
{\mathcal{L}}=  {{\mathcal{L}}_{cls}} + \alpha {\hat{{\mathcal{L}}}_{in}} + \beta{{\mathcal{L}}_{ts}},
\end{equation}
where $\alpha$ and $\beta$ are hyper-parameters.

\section{Experiment}
\subsection{Experimental Setup}
\noindent\textbf{Dataset and Protocols.}
We followed the settings of previous works~\cite{coop,VPT} for both the few-shot learning and domain generalization. In the few-shot setting, we evaluate 11 public datasets including DTD~\cite{dtd}, FGVCAircraft~\cite{aircraft}, Food101~\cite{food101}, ImageNet~\cite{imagenet}, StanfordCars~\cite{cars}, Caltech101~\cite{caltech101}, Flowers102~\cite{flowers}, OxfordPets~\cite{pets}, SUN397~\cite{sun397}, UCF101~\cite{ucf101}, and EuroSAT~\cite{eurosat}. All experiments adopted protocol in CLIP~\cite{clip}, which learns with 1, 2, 4, 8, and 16 labeled samples per class. For the domain generalization setting, we use ImageNet as the source domain and test robustness on target datasets, including ImageNet-A~\cite{imageneta}, ImageNet-R~\cite{imagenetr}, ImageNet-Sketch~\cite{imagenetsk}, and ImageNet-V2~\cite{imagenetv2}.

\begin{table}[h]
    \centering
    \setlength{\belowcaptionskip}{-0.4cm} 
    \small
    \caption{Comparison to state-of-the-art methods for few-shot learning of CLIP-based models, using ViT-B/16 backbone. * indicates the reproduced results where we replace their backbone with VIT-B/16. The best results are in bold.}
    \renewcommand{\arraystretch}{1.2}
    \setlength\tabcolsep{4pt}
    \scalebox{1.00}{
        \begin{tabular}{lccccc}
            \hline
            Method   &        K=1  & K=2 & K=4 & K=8 & K=16  \\   
            \hline
            \textit{Adapter-tuning methods} \\\hdashline 
            $\text{TIP-Adapter}_{\text{ ECCV22}}$~\cite{tip-ada}    & 69.81 & 71.56 & 74.18  & 75.17 & 77.39 \\
            $\text{TIP-Adapter-F}_{\text{ ECCV22}}$~\cite{tip-ada}    & 70.86 &73.10 & 76.04  & 78.81 & 81.27 \\
            $\text{Cross-Modal}_{\text{ CVPR23}}$~\cite{tip-mul}    & 70.75 & 73.29 & 76.79  & 79.05 & 80.75 \\
            $\text{CLAP*}_{\text{ CVPR24}}$~\cite{clap}    & 70.71 & 73.81 & 77.59  & 80.02 & 81.92 \\
            \hline    
            \textit{Prompt-tuning methods}\\ \hdashline
            $\text{CoOP}_{\text{ IJCV22}}$~\cite{coop} & 67.82 & 70.73 & 74.19 & 76.95 & 80.02  \\
            $\text{VPT}_{\text {EECV22}}$~\cite{VPT}  & 66.16  & 68.12 & 70.46 & 74.47 & 76.68  \\
            $\text{CoCoOP}_{\text{ CVPR22}}$~\cite{Cocoop} & 66.79 & 67.65 & 72.21 & 72.96  & 74.92  \\
            $\text{DAPT}_{\text{ ICCV23}}$~\cite{DAPT}  & 61.32 & 69.92  & 75.02 & 79.03 & 81.66 \\
            $\text{MaPLe}_{\text{ CVPR23}}$~\cite{maple}  & 69.40  & 72.51 & 75.62 & 79.13 &81.87  \\
            $\text{LAMM}_{\text{ AAAI24}}$~\cite{maple}  & 68.99  & 73.09 & 75.95 & 78.54 &81.13  \\
            $\text{TCP}_{\text{ CVPR24}}$~\cite{tcp2}    & 70.85 & 73.46 & 76.62  & 78.73 & 81.27 \\
            \rowcolor{gray!30} \textbf{DCA} & \textbf{70.92} & \textbf{74.93} & \textbf{78.95}& \textbf{81.16} & \textbf{83.51} \\
            \hline
        \end{tabular}
    }
    \setlength{\abovecaptionskip}{-0.05cm} 
    \setlength{\belowcaptionskip}{-0.3cm} 
    \label{111}
\end{table}

\begin{table}[h!]
\small
\setlength{\belowcaptionskip}{-0.2cm} 
\centering
\caption{Comparison results of domain generalization.}
\setlength{\belowcaptionskip}{-0.1cm}
\renewcommand{\arraystretch}{1.2}
\setlength{\tabcolsep}{2mm}
\scalebox{0.99}{
    \begin{tabular}{l c c c c c}
 \hline
    \textbf{Method} & \textbf{ImageNet} & \textbf{-V2} & \textbf{-Sketch} & \textbf{-A} & \textbf{-R} \\
 \hline
    CLIP~\cite{clip} & 66.72 & 60.90 & 46.10 & 47.75 & 73.97 \\
    CoOp~\cite{coop} & 71.93 & 64.22 & 47.07 & 48.97 & 74.32 \\
    VPT~\cite{VPT} & 69.31 & 62.36 & 47.72 & 46.20 & 75.81 \\
    MaPLe~\cite{maple} & 70.72 & 64.07 & 49.15 & 50.90 & \textbf{76.98} \\
    DAPT~\cite{DAPT} & 72.20 & 64.93 & 48.30 & 48.74 & 75.75 \\
    TCP~\cite{tcp2} & 71.20 & 64.60 & 49.50 & \textbf{51.20} & 76.73 \\
     \rowcolor{gray!30} \textbf{DCA (Ours)} & \textbf{72.92} & \textbf{66.22} & \textbf{49.53} & 47.86 & 76.53 \\					
    \hline
    \end{tabular}}
    \setlength{\abovecaptionskip}{-0.05cm} 
\setlength{\belowcaptionskip}{-0.3cm} 
\label{tb:gen}
\end{table}

\noindent\textbf{Compare Methods.}
We compare several classical prompt learning methods, including CoOP~\cite{coop} and VPT~\cite{VPT}. We also compare our approach with SOTA methods such as MaPLe~\cite{maple}, DAPT~\cite{DAPT}, and TCP~\cite{tcp2}. Moreover, we compare with several SOTA adapter-tuning methods, including Tip-Adapter~\cite{tip-ada}, Cross-Modal~\cite{tip-mul}, and CLAP~\cite{clap}.

\noindent\textbf{Implementation Details.} Our implementation is based on the ViT-B/16 variant of the CLIP model. We set the length of the visual prompt to 16 and configured the text prompt lengths to range from 3 to 6. The number of text prompts $M$ is set to 4. We evaluate all experiments with \textbf{three} seeds and report the average value. More details are provided in the \textbf{appendix}.

\begin{table*}[h]
\setlength{\belowcaptionskip}{-0.2cm} 
\centering
\small
\caption{Ablation study of proposed DCA. We adopt CA, ${\hat{{\mathcal{L}}}_{in}}$, DA and ${\mathcal{L}}_{ts}$ to denote covariance-aware, Mahalanobis intra-class loss diversity-aware and text separation loss.}
\renewcommand{\arraystretch}{1.2}
\setlength{\tabcolsep}{2mm}
\scalebox{1}{
\begin{tabular}{cccc | ccccccccc}
   \hline
  CA & ${\hat{{\mathcal{L}}}_{in}}$ & DA & ${\mathcal{L}}_{ts}$ &  DTD  & FGVCAircraft & StanfordCars  & Flowers102 & UCF101 & Food101 & EuroSAT  \\ \hline
   & & &             &  60.91     & 29.89             &   72.01      &  90.95          &    79.12        &  83.12       &  84.33                 \\
  \checkmark& & &            &  63.81     &  31.11            &  72.93       &    91.85        &   79.81         &  84.13      & 87.53                   \\
  \checkmark& \checkmark & &           & 64.39     & 33.94           &   73.84      &   91.95         &  80.62          &  84.72       &  88.72                 \\ 
  \checkmark& & \checkmark &           &  64.94     & 36.65           &   74.45      &   93.88         &  80.92          &  84.55       &  88.92                   \\ 
   \checkmark& \checkmark & \checkmark&           &  65.54     & 38.78           &   74.98      &   93.96         &  81.15          &  84.74       &  88.95                  \\ 
  \checkmark& \checkmark & \checkmark& \checkmark       &  \textbf{67.61}     &  \textbf{39.39}            &  \textbf{75.96}       &   \textbf{94.29}         &   \textbf{81.66}         &  \textbf{85.12}       &  \textbf{89.98}                     \\ \hline

\end{tabular}}
\label{tb:ablation}
\end{table*}

\begin{figure*}[htb!]
\setlength{\abovecaptionskip}{-0.1cm} 
\setlength{\belowcaptionskip}{-0.2cm} 
\begin{center}
    \includegraphics[width=\textwidth]{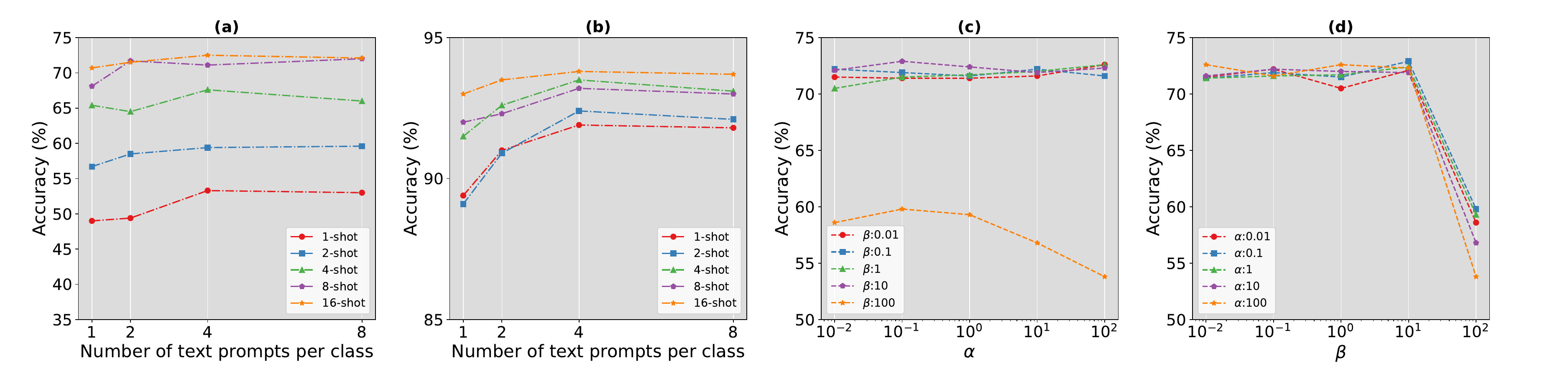}
\end{center}
\caption{(a,b) The influence of text prompt number $M$. (c,d) Sensitivity analysis of hyper-parameters.}
\label{fig:sensitive}
\end{figure*}

\subsection{Few-shot Learning}
We present the comparative results with SOTA methods in Fig.~\ref{fig:coco} and Table~\ref{111}. We summarize the result as follows:
\begin{itemize}
    \item For most datasets, all tuning methods outperform zero-shot CLIP. Our DCA demonstrates a more significant performance improvement, achieving a maximum boost of \textbf{18.16\%} at 16 shots.
    \item For the adapter-tuning methods, our approach achieves a performance improvement with fewer additional parameters and higher efficiency, with a maximum improvement of \textbf{1.59\%} at 16 shots.
    \item Compared to classical methods such as CoCo and VPT, the DCA consistently improves all shots. For instance, DCA provides an average absolute gain ranging from \textbf{4.76\%} to \textbf{8.46\%} from 1 to 16 shots. For CoOp, we also achieved an average improvement of \textbf{3.95\%}, with a maximum improvement of \textbf{4.76\%} at 4 shots.
    \item  For the SOTA methods, such as MaPLe, DAPT, and TCP, our DCA also achieves significant improvements, especially with increases of at least \textbf{2.33\%}, \textbf{2.06\%}, and \textbf{1.64\%} from 4 to 16 shots, respectively. Notably, our method maintains \textit{stable performance} across different shots, avoiding the severe performance drop observed with the DAPT at 1 shot, demonstrating its robustness.
\end{itemize}

\subsection{Domain Generalization}
We evaluate the generalization of DCA by comparing it with zero-shot CLIP and prompt learning methods in a domain-generalization setting. We use ImageNet as the source dataset, with prompts trained on 16 samples, and four datasets as the target datasets. The overall results are shown in Table~\ref{tb:gen} and we can summarize that:
\begin{itemize}
    \item DCA achieves significant performance improvements on unseen data compared to zero-shot CLIP, with a maximum increase of \textbf{5.32\%} on ImageNet-V2.
    \item For DAPT, which also focuses on few-shot tasks like ours, our method demonstrates superior performance across most datasets with a significant accuracy gain, showcasing better generalization.
    \item For methods that focus more on generalization, such as MaPLe and TCP, accuracy decreases on ImageNet-A and ImageNet-R, respectively. In contrast, on the remaining datasets, especially ImageNet-V2, there is up to a \textbf{1.62\%} improvement. 
\end{itemize}

\subsection{Ablation Study}

\begin{table*}[t]
\small
\setlength{\belowcaptionskip}{-0.3cm} 
\centering
\caption{The effectiveness analysis of covariance modeling in visual-language models. }
\renewcommand{\arraystretch}{1.2}
\setlength{\tabcolsep}{1.5mm}
\scalebox{1}{
\begin{tabular}{lccccc|ccccc|ccccc}
 \hline
\multirow{2}{*}{Method} & \multicolumn{5}{c|}{FGVCAircraft} & \multicolumn{5}{c|}{DTD} & \multicolumn{5}{c}{Caltech101} \\ \cline{2-16} 
                        & 1 & 2 & 4 & 8 & 16 & 1 & 2 & 4 & 8 & 16 & 1 & 2 & 4 & 8 & 16 \\ \hline
Baseline+$d_{cos}$            & 10.28  & 15.22  & 29.96  & 40.82  & 46.27   & 9.28  & 45.32  & 60.93  & 65.09  & 70.52   & 91.78  & 92.68  & 92.69  & 94.78  & 95.75   \\
\textbf{Baseline+$\boldsymbol{d_m}$}            & \textbf{16.62}  & \textbf{20.88}  & \textbf{31.11}  & \textbf{41.72}  & \textbf{46.98}   & \textbf{10.98}  & \textbf{57.18}  & \textbf{63.81}  & \textbf{69.46}  & \textbf{70.73}   & \textbf{92.32}  & \textbf{93.84}  & \textbf{94.16}  & \textbf{95.21}  & \textbf{95.84}  \\  \hline
DAPT+$d_{cos}$                 & 4.32  & 20.90  & 32.88  & 41.39  & 46.29   & 20.61  & 54.55  & 60.68  & 66.38  & 71.28   & 92.18  & 93.28  & 94.29  & 94.77  & 95.76  \\
 \textbf{DAPT+$\boldsymbol{d_m}$}                & \textbf{16.33}  & \textbf{21.64}  & \textbf{33.94}  & \textbf{42.62}  & \textbf{49.08}   & \textbf{21.88}  & \textbf{57.22}  & \textbf{62.41}  & \textbf{69.63}  & \textbf{71.58}   & \textbf{92.94}  & \textbf{93.84}  & \textbf{94.43}  & \textbf{95.21}  & \textbf{96.18}   \\  \hline
DCA+$d_{cos}$                 & 26.11  & 31.62  & 36.92  & 44.35  & 51.12  & 47.93  & 53.45  & 63.12  & 68.32  & 72.14   & 89.92  & 92.31  & 92.87  & 94.46  & 95.38  \\
 \textbf{DCA+$\boldsymbol{d_m}$}                 & \textbf{30.33}  & \textbf{33.64}  & \textbf{39.39}  & \textbf{45.82}  & \textbf{52.38}   & \textbf{53.32}  & \textbf{59.42}  & \textbf{67.61}  & \textbf{72.52}  & \textbf{74.25}   & \textbf{94.34}  & \textbf{94.42}  & \textbf{94.71}  & \textbf{95.81}  & \textbf{96.34}   \\ \hline

\end{tabular}}
\label{tb:anal}
\end{table*}

\noindent\textbf{The Effectiveness of Each Component.} To demonstrate how each component of the proposed DCA contributes to performance improvements, we conducted ablation studies of 4-shot learning on 7 datasets. Table~~\ref{tb:ablation} shows the results and we can observe that:
\begin{itemize}
   \item All components in DCA consistently improve the baseline performance without any conflicts between them. The improvements are particularly notable on challenging datasets, \textit{e.g.} DTD and FGVCircraft, where the cumulative improvements are \textbf{6.70\%} and \textbf{9.50\%}, respectively.
   \item Compared to the baseline, the CA method significantly improves across multiple datasets, with a maximum gain of \textbf{3.20\%} on EuroSAT. This validates our hypothesis that feature spaces in few-shot learning tasks benefit more from anisotropic distances, \textit{i.e.} Mahalanobis distance. The DA method consistently enhances CA’s performance across all datasets, achieving up to \textbf{5.54\%} improvement on FGVCAicraft, showing that multi-centered covariance modeling enhances decision boundaries.
    \item The Mahalanobis intra-class and text separation losses respectively increase performance by an average of \textbf{1.0\%} and \textbf{0.95\%}, demonstrating that these loss functions can further optimize the feature space, enhancing the effectiveness of both CA and DA methods.
\end{itemize}

\begin{figure}[t!]
 \setlength{\abovecaptionskip}{-0.1cm} 
\setlength{\belowcaptionskip}{-0.3cm} 
\begin{center}
    \includegraphics[width=0.5\textwidth]{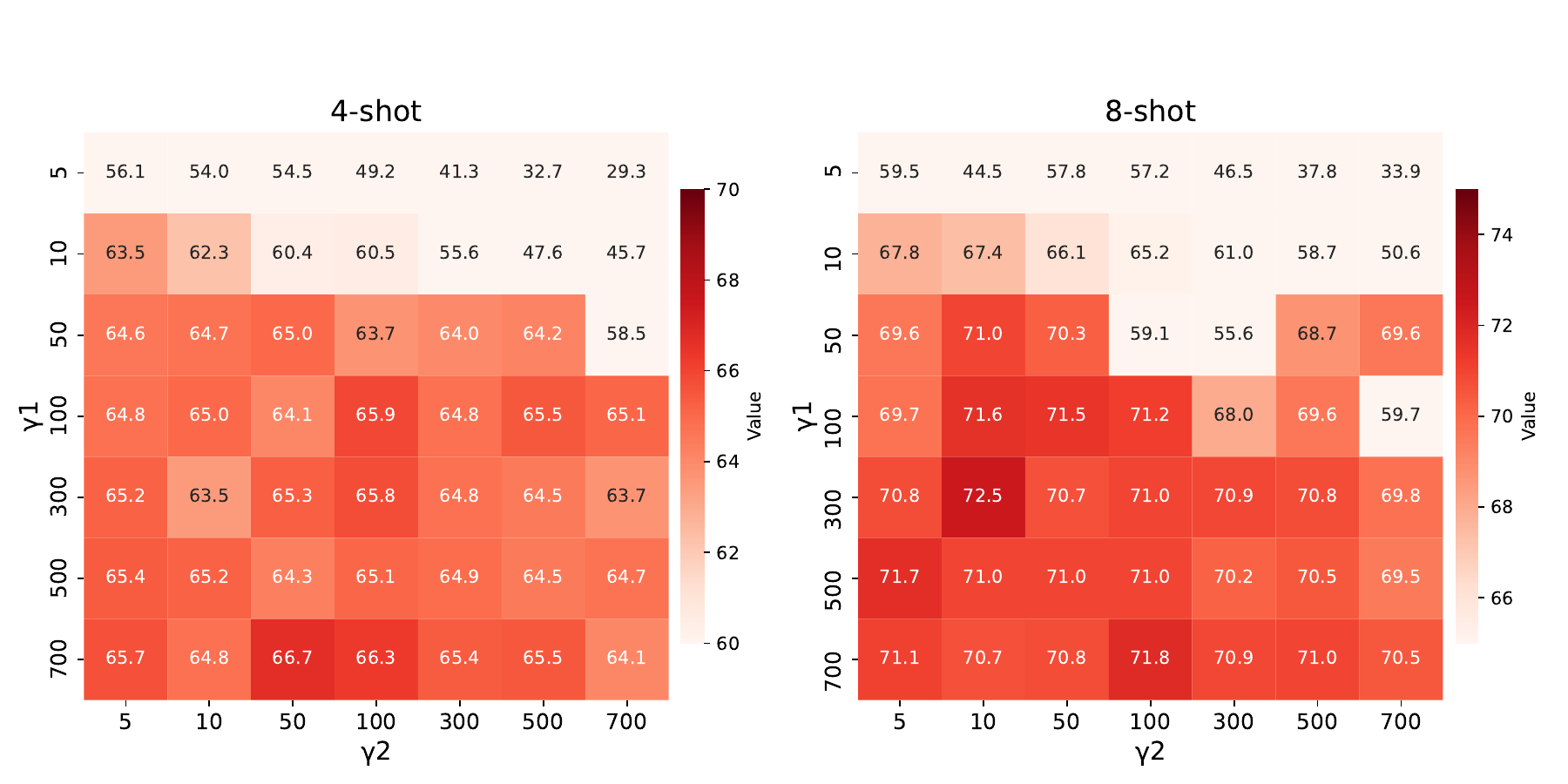}
\end{center}
 \setlength{\abovecaptionskip}{-0.1cm} 
\caption{Impact of covariance shrinkage parameter $\gamma _1$ and  $\gamma _2$.}
 \setlength{\abovecaptionskip}{-0.1cm} 
\label{fig:heatmap_plots}
\end{figure}

\noindent\textbf{The Ablation Study of Text Prompt Number.}
As discussed in Sec.~\ref{diversity}, $M$ represents the number of text prompts. We conduct an ablation study to evaluate the impact of the number of text prompts per class. Fig.~\ref{fig:sensitive} (a) and (b) show the results on the DTD and OxfordPet. We observed that two datasets showed consistent results: (1) performance increases as the number of text prompts increases, indicating that more text prompts can help learn comprehensive visual information. (2) Notably, performance peaks at $M$=$4$ and begins to decline slightly as $M$ increases, suggesting that too many prompts can lead to information redundancy, which is detrimental to optimization.

\noindent\textbf{The Sensitivity of Hyper-parameter.}
(1) The hyper-parameters $\alpha$ and $\beta$ adjust the strength of the Mahalanobis intra-class loss and the text diversity loss. We conducted a sensitivity study on the DTD dataset, where we varied the values of $\alpha$ and $\beta$ within the reasonable range of $\{10^{-2}, 10^{-1}, 1, 10^{1}, 10^{2}\}$, respectively. As shown in Fig.~\ref{fig:sensitive} (c) and (d), except for $\beta$ being $10^{2}$, the performance of DCA varies only slightly and shows consistent performance across a wide range of hyper-parameters. These indicate the robustness of our method.

(2) We also analyzed the effects of the covariance shrinkage parameters $\gamma_1$ and $\gamma_2$ in the few-shot settings. Take DTD as an example, as shown in Fig.~\ref{fig:heatmap_plots}. Our observations indicate that the model performance remained stable for both the 4-shot and 8-shot settings and achieved optimal results when $\gamma_1 \in [100, 700]$ and $\gamma_2 \in [5, 500]$. Similarly, we can obtain the optimal values of $\gamma_1$ and $\gamma_2$ for the other 10 datasets under different shot settings.

Due to the different characteristics of each dataset and each shot setting, there may be different optimal parameter values. Therefore, we provide the optimal hyper-parameters for each dataset and each shot setting in the \textbf{appendix}.

\subsection{Analysis}

\noindent\textbf{The Effectiveness of Covariance Modeling in VLMs.}
Compared to other methods using cosine distance, we use Mahalanobis distance to measure the distance of text and image features. To further verify the effectiveness of the CA method, we applied it to the baseline and SOTA method DAPT across three datasets. Table~\ref{tb:anal} shows the experimental results for 1$\sim$16 shots. We can observe that covariance modeling consistently improves performance across the three datasets, especially when the training samples are limited, such as in the 1-shot setting. For example, it improves the baseline and DAPT by \textbf{6.34\%} and \textbf{12.01\%}, respectively, on FGVCAircraft. In addition, compared with cosine distance, using Mahalanobis distance also has a significant effect on our method.

\begin{table}[t!]
\small
 \setlength{\abovecaptionskip}{-0.1cm} 
\caption{The effectiveness analysis of covariance modeling in intra-class loss.}
\begin{center}
\renewcommand{\arraystretch}{1.2}
\setlength{\tabcolsep}{2mm}
\scalebox{1.0}{
\begin{tabular}{lcc}\hline
\multirow{2}{*}{Method} & \multicolumn{2}{c}{Average over 11 datasets} \\ \cline{2-3} 
                        & \textbf{4 shots} & \textbf{16 shots} \\ \hline
$\mathcal{L}_{in}$-L1            & 77.96 ($\downarrow$ 0.99) & 82.62 ($\downarrow$ 0.89) \\
$\mathcal{L}_{in}$-L2            & 78.02 ($\downarrow$ 0.81) & 82.70 ($\downarrow$ 0.72) \\
 $\mathcal{L}_{in}$-$d_m$ (ours)     & \textbf{78.95} & \textbf{83.51} \\

\hline
\end{tabular}}
\end{center}
\vspace{-0.5cm}
\label{tb:comparison}
\end{table}

\noindent\textbf{The Effectiveness of Mahalanobis Intra-class Loss.} We use Mahalanobis distance instead of Euclidean distance to achieve intra-class contraction. To demonstrate the performance improvement of the loss function with Mahalanobis distance, we conducted analytical experiments on 11 datasets. As shown in Table 4, using Mahalanobis distance results in performance improvements for both 4-shot and 16-shot settings, with maximum increases of \textbf{0.99\%} and \textbf{0.81\%} over L1 and L2 distance, respectively. This demonstrates that covariance modeling can adjust the loss function based on the dataset's distribution, helping to prevent overfitting.

\section {Conclusion and Limitations}
We proposed a prompt learning framework called DCA for VLMs. By modeling the covariance in the visual feature space, using Mahalanobis distance to measure between text and visual modalities, and learning multiple diverse and information-rich soft prompts, our method significantly enhances prompt model performance without substantially increasing computational overhead. Meanwhile, we theoretically prove the rationality of our method. Although the proposed method significantly improves performance on few-shot tasks, it still faces challenges in model generalization tasks, such as domain generalization. Therefore, maintaining model generalization to unseen tasks during downstream training is a promising direction for future research.

\input{suppl}

{
    \small
    \bibliographystyle{ieeenat_fullname}
    \bibliography{main}
}

\end{document}

%% file: suppl.tex
\section{APPENDIX}
\subsection{The Proof of Bayesian Classifier}

In the main paper, based on the principle of maximization and Bayes’ theorem~\cite{goswami2024fecam,kumar2022gdc}, we obtained the following equation:
\begin{equation}
\label{basi}
\arg \max_{y} P(Y \mid X) = \arg \max_{y} \log P(X \mid Y).
\end{equation}
Next, we can present a detailed proof of this equation. Specifically, applying Bayes' theorem, we have:
\begin{equation}
    P(Y \mid X) = \frac{P(X \mid Y)P(Y)}{P(X)},
\end{equation}
where $P(Y \mid X)$ represents the posterior probability of class $Y$ given $X$,
$P(X \mid Y)$ denotes the likelihood of $X$ given class $Y$,
$P(Y)$ refers to the prior probability of class $Y$,
and $P(X)$ corresponds to the marginal probability of $X$. Since the marginal probability $P(X)$ remains constant for all $y$, it can be disregarded when maximizing $P(Y \mid X)$:
\begin{equation}
\begin{aligned}
\arg \max_{y} P(Y \mid X) &= \arg \max_{y} \frac{P(X \mid Y)P(Y)}{P(X)} \\
&= \arg \max_{y} P(X \mid Y)P(Y). \\
\end{aligned}
\end{equation}
Assuming the prior probabilities $P(Y)$ are identical across all classes, we take the logarithm of the expression:
\begin{equation}
\begin{aligned}
\arg \max_{y} &\log \left( P(X \mid Y)P(Y) \right) \\
 &=\arg \max_{y} \left( \log P(X \mid Y) + \log P(Y) \right) \\
&= \arg \max_{y} \log P(X \mid Y),
\end{aligned}
\end{equation}
where based on Eq. (3) and Eq. (4), we can derive Eq. (1), thus concluding the proof.

Given the anisotropic Gaussian distribution assumption  \(\mathcal{N}(\mu_y, \Sigma_y)\) to model the feature distribution of class \( y \), and , we have:
\begin{equation}
\begin{aligned}
    &P(x \mid c = y) = \\ 
    &\frac{1}{(2\pi)^{D/2} |\Sigma_y|^{1/2}} \exp \left( -\frac{1}{2} (x - \mu_y)^T \Sigma_y^{-1} (x - \mu_y) \right),
\end{aligned}
\end{equation}
where $D$ is the dimensional of $x$, $\mu_y=\frac{1}{N_{i}}\sum_{n=1}^{N_{i}}x_i$ is the average feature vector of all feature of class $y$. $x, \mu_y \in R^D$ and $\Sigma^{-1}_y$ is an arbitrary positive definite matrix. Then, we substitute Eq. (5) into Eq. (1) to obtain the following formula:
\begin{equation}
\begin{aligned}
&\arg \max_{y} P(Y \mid X) = \arg \max_{y} \log P(X \mid Y) \\
&= \arg \max_{y} -\frac{1}{2} \left[ (x - \mu_y)^T \Sigma_y^{-1} (x - \mu_y) + \log |\Sigma_y| + D \log (2\pi) \right] \\
&= \arg \min_{y} \, (x - \mu_y)^T \Sigma^{-1}_y (x - \mu_y).
\end{aligned}
\end{equation}

Therefore, based on the above derivation, we transform the Bayesian classification problem into a problem of minimizing the Mahalanobis distance.

\subsection{Implementation Details}

\noindent \textbf{Model and Configuration.}  Our implementation is based on the ViT-B/16 variant of the CLIP model. We set the length of the visual prompt to 16 and configured the text prompt lengths to range from 3 to 6. The number of each class text prompts $M$ is set to 4.  The text prompts are initialized with the following phrases: “A photo of  [CLASS]”, “A drawing of a  [CLASS]”, “This is a photo of [CLASS]”, and “This is a photo of a [CLASS]”.

\noindent \textbf{Covariance Matrix Shrinkage.} For stability in covariance matrix estimation, we apply shrinkage with parameters $\gamma_1 $ and $\gamma_2 $ for different shot settings in Table~\ref{tb:1}. This ensures that the covariance matrix remains robust throughout the training process.

\noindent \textbf{Training Protocol.} Following the comparative approaches~(\textit{e.g.} CoOp and DAPT), we train our model with different numbers of epochs depending on the few-shot setting. Specifically, we use 80 epochs for the 1-shot setting, 100 epochs for the 2-shot and 4-shot settings, and 200 epochs for the 8-shot and 16-shot settings.

\noindent \textbf{Optimizer and Learning Rate Schedule.} We employ the SGD optimizer for training, with the learning rate adjusted according to a cosine schedule. For warmup, we apply 5 epochs with a constant learning rate of 1e-5.

\noindent \textbf{Batch Size.} Similar to other approaches~\cite{DAPT,coop}, the batch size varies according to the dataset size. For smaller datasets such as FGVCAircraft, OxfordFlowers, and StanfordCars, the batch size is set to 32. For larger datasets like Imagenet and SUN397, the batch size is set to 64.

\noindent \textbf{Hardware and Platform.} We implemented our method with the PyTorch platform and trained on 2 RTX 3090 24G GPUs.

\begin{figure}[t!]
 \setlength{\abovecaptionskip}{-0.1cm} 
\setlength{\belowcaptionskip}{-0.3cm} 
\begin{center}
    \includegraphics[width=0.48\textwidth]{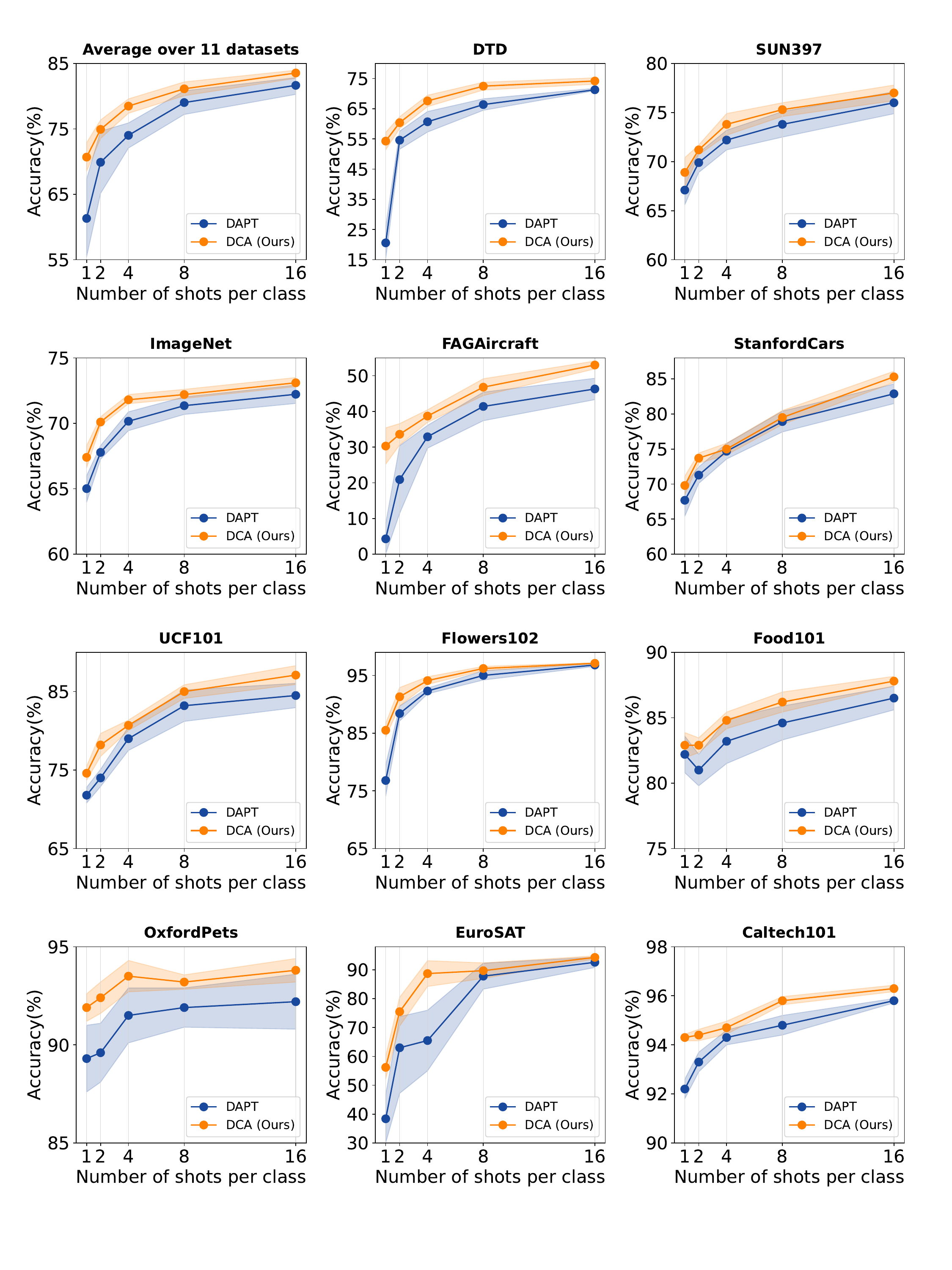}
\end{center}
 \setlength{\abovecaptionskip}{-0.7cm} 
\caption{Robustness Analysis. Our method (DCA) is compared with the SOTA method (DAPT~\cite{DAPT}) across eleven datasets with different random seeds. The results show that DCA achieves better average performance and exhibits lower variance, indicating superior stability and robustness under different random seed settings.}

\label{fig:Robustness}
\end{figure}

\noindent \textbf{Seed and Averaging.} To ensure the reliability of our results, all experiments are conducted with \textbf{three} different seeds(1/2/3), and the results are averaged across these runs, which is shown in Figure~\ref{fig:Robustness}.

\subsection{More Compare Results}
In this section, we provide additional comparative methods and detailed comparative results for each experimental setup in Table~\ref{2}. 

\subsection{Training and Testing Loop}
We provide a detailed training and testing loop to help understand our DCA method in Algorithm\ref{1111}.

\begin{algorithm}
\caption{The training and testing loop of the proposed DCA method for vision-language models}
\label{1111}
\begin{algorithmic}[1]
\Require Pre-trained visual encoder $z(\cdot)$, text encoder $g(\cdot)$, training dataset $D$, number of classes $N$, number of text prompts per class $M$, hyper-parameters $\alpha, \beta$, learning rate $\eta$
\Ensure Optimized text prompts $\mathbf{w}$ and visual prompt $\mathbf{v}$

\State \textbf{Initialization:} Initialize prompts $\{ \mathbf{w}^1, \ldots, \mathbf{w}^M \}$ for each class, visual prompt $\mathbf{v}$, and covariance matrix $\Sigma_y = \mathbf{I}$ for each class $y$.

\For{each class $y$ in $N$}
    \State Compute class center $\mu_y = \frac{1}{|D_y|} \sum_{x \in D_y} z(x)$
\EndFor

\While{not converged}
    \For{each mini-batch $B = \{(x_i, y_i)\}$ from $D$}
        \For{each image $x_i$ with label $y_i$ in $B$}
            \State Compute visual feature $f_i = z(x_i)$
            \For{each $p^m$ in $\{ p^1, \ldots, p^M \}$}
                \State Compute text feature $u^m = g(p^m)$
                \State Calculate distance $d_e(f_i, u^m)$ (Eq.7)
                \State Compute prediction probability $p(c=y_i \mid x_i)$ (Eq.8)
                \State Compute cross-entropy loss $\mathcal{L}_{cls}^m = -\log p(c=y_i \mid x_i)$ (Eq.9)
            \EndFor
            \State Sum losses from all $M$ prompts to obtain $\mathcal{L}_{cls} = \sum_{m=1}^{M} \mathcal{L}_{cls}^m$
        \EndFor
        \State Compute intra-class loss $\hat{\mathcal{L}}_{in}$ with covariance $\hat{\Sigma}_{s_{y_i}}$ (Eq.11)
        \State Compute text separation loss $\mathcal{L}_{ts}$ (Eq.12)
        \State \textbf{Total Loss:} $\mathcal{L} = \mathcal{L}_{cls} + \alpha \hat{\mathcal{L}}_{in} + \beta \mathcal{L}_{ts}$ (Eq.13)
        \State \textbf{SGD Update:} $\mathbf{w} \leftarrow \mathbf{w} - \eta \nabla_{\mathbf{w}} \mathcal{L}$, $\mathbf{v} \leftarrow \mathbf{v} - \eta \nabla_{\mathbf{v}} \mathcal{L}$
    \EndFor
\EndWhile

\State \textbf{Testing Phase:}
\For{each test image $x$}
    \State Compute visual feature $f = z(x)$
    \State Compute covariance matrix $\Sigma_y$ and apply shrinkage to obtain $\Sigma_{s_y}$
    \State Compute Mahalanobis distance $d_m(f, u_y)$ using $\hat{\Sigma}_{s_{y}}$
    \State Predict class label based on smallest Mahalanobis distance
\EndFor

\end{algorithmic}
\end{algorithm}


\begin{table*}[t]

\centering
\scriptsize 
\caption{The hyper-parameters $\gamma_1$ and $\gamma_2$ for different shot settings in few-shot learning.}
\renewcommand{\arraystretch}{1.2} 
\setlength{\tabcolsep}{5mm} 
\scalebox{1}{ 
\begin{tabular}{lccccc}
\toprule
\multirow{2}{*}{\textbf{Hyper-parameters}} & \multicolumn{5}{c}{\textbf{Shot Settings}} \\ \cmidrule(lr){2-6}
 & \textbf{1-shot} & \textbf{2-shot} & \textbf{4-shot} & \textbf{8-shot} & \textbf{16-shot} \\
\midrule
$\gamma_1$ & 500 & 500 & 600 & 500 & 500 \\
$\gamma_2$ & 300 & 300 & 100 & 100 & 500 \\
\bottomrule
\end{tabular}}

\label{tb:1}
\end{table*}

\begin{table*}[htbp]
\setlength{\abovecaptionskip}{0.1cm} 
\setlength{\belowcaptionskip}{0.3cm} 
\setlength{\tabcolsep}{1.5mm}
\scriptsize
\centering
\caption{Per-dataset performance comparison of DCA with various methods in few-shot setting. * denotes the performance obtained by our re-implementation. The best results are in \textbf{bold} and the second-best results are \underline{underlined}.}
\label{tab:base_2_novel_app}
\begin{tabular}{lccccccccccc}
\hline
\textbf{Dataset} & \textbf{Set} & \begin{tabular}[c]{@{}c@{}}CLIP\\ \end{tabular} & \begin{tabular}[c]{@{}c@{}}LP+CLIP\\ \end{tabular} & \begin{tabular}[c]{@{}c@{}}CoOp\\ (\tiny IJCV22)\end{tabular} & \begin{tabular}[c]{@{}c@{}}VPT\\ (\tiny ECCV22)\end{tabular} & \begin{tabular}[c]{@{}c@{}}CoCoOp\\ (\tiny CVPR22)\end{tabular} & \begin{tabular}[c]{@{}c@{}}DAPT\\ (\tiny ICCV23)\end{tabular} & \begin{tabular}[c]{@{}c@{}}MaPLe\\ (\tiny CVPR23)\end{tabular} & \begin{tabular}[c]{@{}c@{}}LAMM\\ (\tiny AAAI24)\end{tabular}& \begin{tabular}[c]{@{}c@{}}TCP*\\ (\tiny CVPR24)\end{tabular} & \textbf{Ours} \\ \hline

\multirow{5}{*}{\textbf{Average}} 
& 1 shot  & 65.35 & 45.83 & 67.82 & 66.16 & 66.79 & 61.32 & 69.4 & 68.97  & \underline{70.85} &\textbf{ 70.92} \\
& 2 shots & 65.35 & 57.98 & 70.73 & 68.12 & 67.65 & 69.92 & 72.51 & 73.22 &\underline{ 73.46} & \textbf{74.93} \\
& 4 shots & 65.35 & 68.01 & 74.19 & 70.46 & 72.21 & 75.02 & 75.62 & 75.89 &\underline{ 76.62} & \textbf{78.95} \\
& 8 shots & 65.35 & 74.47 & 76.95 & 74.47 & 72.96 & 79.03 &\underline{ 79.13} & 78.47 & 78.73 & \textbf{81.16} \\
& 16 shots & 65.35 & 78.79 & 80.02 & 76.68 & 74.92 & 81.66 &\underline{ 81.87} & 81.07 & 81.27 & \textbf{83.51} \\\hline
\multirow{5}{*}{\textbf{ImageNet}} & 1 shot                      & 66.67                                                                         & 32.13                                                                         & 66.33                                                                        & 66.91                                                                          & \textbf{69.43}                                                                        & 65.01                                                                         & 62.47  & 67.22                                                                     & \underline{ 68.82 }                                                                                                       & 67.43 \\
                                  & 2 shots                     & 66.67                                                                         & 44.88                                                                         & 67.07                                                                        & 67.92                                                                          & \underline{ 69.78 }                                                                       & 67.78                                                                         & 65.13  & 68.67                                                                      & 69.52                                                                                                       & \textbf{ 70.14} \\
                                  & 4 shots                     & 66.67                                                                         & 54.85                                                                         & 68.73                                                                        & 68.73                                                                          & 70.39                                                                        & 70.16                                                                         & 67.72        & 69.88                                                                & \underline{ 70.53}                                                                                                      & \textbf{ 71.82} \\
                                  & 8 shots                     & 66.67                                                                         & 62.23                                                                         & 70.63                                                                        & 68.97                                                                          & 70.65                                                                        & 71.35                                                                         & 70.32       & 71.32                                                                 & \underline{ 71.23}                                                                                                       & \textbf{ 72.23} \\
                                  & 16 shots                     & 66.67                                                                         & 67.31                                                                         & 71.87                                                                        & 69.36                                                                          & 70.83                                                                        & 72.22                                                                         & 72.33            & \underline{72.71}                                                            &  72.4                                                                                                       & \textbf{ 73.13} \\\hline

\multirow{5}{*}{\textbf{Caltech101}}    & 1 shot                      & 93.32                                                                         & 79.88                                                                         & 92.63                                                                         & 92.58                                                                        & 93.83                                                                      & 92.22                                                                         & 92.57     & 93.39                                                                  & \underline{ 94.13}                                                                                                        & \textbf{ 94.32} \\
                                  & 2 shots                     & 93.32                                                                         & 89.01                                                                         & 93.07                                                                        & 93.62                                                                         & \textbf{ 94.82}                                                                      & 93.3                                                                         & 93.97        & 93.90                                                               & 94.42                                                                                                       & \underline{ 94.43} \\
                                  & 4 shots                     & 93.32                                                                         & 92.05                                                                         & 94.41                                                                         & 94.61                                                                         & \underline{ 94.98}                                                                      & 94.32                                                                         & 94.43        & 94.81                                                               & \textbf{ 95.63 }                                                                                                      & 94.73 \\
                                  & 8 shots                     & 93.32                                                                         & 93.41                                                                         & 93.37                                                                        & 94.82                                                                         & 95.04                                                                      & 94.81                                                                         & 95.21         & 95.79                                                               & \underline{ 95.32}                                                                                                       & \textbf{ 95.84} \\
                                  & 16 shots                    & 93.32                                                                         & 95.43                                                                         & 95.57                                                                        & 95.36                                                                        & 95.16                                                                      & 95.81                                                                         & \underline{ 96.02}                                                                & 96.89        & 95.83                                                                                                       & \textbf{ 96.34} \\ \hline

\multirow{5}{*}{\textbf{StanfordCars}}    & 1 shot & 65.54 & 35.66 & 67.43 & 64.73 & 67.22 & \underline{ 67.69} & 66.62   &67.34     & 66.43& \textbf{ 69.84} \\
& 2 shots & 65.54 & 50.28 & 70.51 & 66.52 & 68.37 & 71.28 &  71.63 &\textbf{73.11}     & 70.74 & \underline{ 72.72} \\
& 4 shots & 65.54 & 63.38 & 74.47 & 67.31 & 69.39 & 74.69 & { 75.32}  & \textbf{77.58}       & { 75.73} & \underline{75.96} \\
& 8 shots & 65.54 & 73.67 & 79.31 & 69.62 & 70.44 & 78.92 & { 79.47} & \textbf{81.29}    & 79.12 & \underline{ 79.58} \\
& 16 shots & 65.54 & 80.44 & 83.07 & 70.52 & 71.57 & 82.88 & 83.57 & \underline{85.07}    & { 84.02} & \textbf{ 85.38} \\ \hline
\multirow{5}{*}{\textbf{Flowers102}} & 1 shot                      & 70.73                                                                         & 69.74                                                                         & 77.53                                                                        & 68.27                                                                          & 69.74                                                                        & 76.81                                                                         & 83.31        &84.49                                                                    & \textbf{ 86.41}                                                                                                        & \underline{ 85.51} \\
                                  & 2 shots                     & 70.73                                                                         & 85.07                                                                         & 87.33                                                                        & 73.81                                                                          & 85.07                                                                        & 88.41                                                                         & 88.93             & \textbf{91.72}                                                               & { 90.81}                                                                                                       & \underline{ 91.31} \\
                                  & 4 shots                        & 70.73                                                                         & 92.02                                                                         & 92.17                                                                        & 77.83                                                                          & 92.02                                                                        & 92.31                                                                         & 92.67         & 93.23                                                                   & \textbf{ 94.91}                                                                                                       & \underline{ 94.21} \\
                                  
                                  & 8 shots                       & 70.73                                                                         & 96.11                                                                         & 94.97                                                                        & 87.62                                                                          & 96.11                                                                        & 95.01                                                                         & 95.81        & 95.89                                                                    & \textbf{ 96.31}                                                                                                       & \underline{ 96.21} \\
                                  
                                   & 16 shots                        & 70.73                                                                         & 97.35                                                                         & 97.07                                                                        & 91.42                                                                          & \underline{ 97.37}                                                                        & 96.81                                                                         & 97.01                   &  \textbf{97.38}                                                        & { 97.31}                                                                                                       & 97.11 \\\hline
\multirow{5}{*}{\textbf{FGVCAircraft}} & 1 shot & 24.31 & 19.61 & 21.37 & 23.26 & 12.68 & 4.32 & 26.73 &27.72     & \underline{ 29.71} & \textbf{ 30.33} \\
& 2 shots & 24.31 & 26.41 & 26.2 & 24.55 & 15.06 & 20.92 & 30.92 &32.13     & \underline{ 32.31} & \textbf{ 33.63} \\
& 4 shots & 24.31 & 32.33 & 30.83 & 27.38 & 24.79 & 32.91 & 34.87 & 35.31    & \underline{ 36.22} & \textbf{ 39.39} \\
& 8 shots & 24.31 & 39.35 & 39.02 & 30.86 & 26.61 & 41.41 & \underline{ 42.02} & 40.67    & 40.73 & \textbf{ 45.83} \\
& 16 shots & 24.31 & 45.36 & 43.42 & 34.98 & 31.21 & 46.32 & \underline{ 48.41} & 45.38    & 44.53 & \textbf{ 52.38} \\ \hline
\multirow{5}{*}{\textbf{DTD}}     & 1 shot                      & 44.61                                                                         & 34.59                                                                         & 50.23                                                                        & 45.19                                                                          & 48.54                                                                        & 20.61                                                                         & 52.13           &51.69                                                                 & \textbf{ 53.41}                                                                                                        & \underline{ 53.31} \\
                                  & 2 shots                     & 44.61                                                                         & 40.76                                                                         & 53.60                                                                         & 46.21                                                                          & 52.17                                                                        & 54.62                                                                         & 55.51             &60.01                                                                & \underline{ 57.01}                                                                                                       & \textbf{ 59.43} \\
                                  & 4 shots                        & 44.61                                                                         & 55.71                                                                         & 58.73                                                                         & 50.66                                                                          & 55.04                                                                        & 60.74                                                                         & 61.03           & 64.99                                                                 & \underline{ 62.53 }                                                                                                      & \textbf{ 67.62} \\
                                  
                                  & 8 shots                       & 44.61                                                                         & 63.46                                                                         & 64.77                                                                        & 56.51                                                                          & 58.89                                                                        & 66.42                                                                         & 66.52            & 66.91                                                                & \underline{ 68.43 }                                                                                                      & \textbf{ 72.52} \\
                                  
                                   & 16 shots                        & 44.61                                                                         & 69.96                                                                         & 69.87                                                                        & 61.47                                                                          & 63.04                                                                        & 71.35                                                                         & 71.33                                                               & \underline{72.19}             & { 72.03 }                                                                                                      & \textbf{ 74.25} \\\hline
\multirow{5}{*}{\textbf{SUN397}}  
& 1 shot & 62.61 & 41.58 & 66.77 & 65.22 & 68.33 & 67.13 & 64.77 &66.69     & \textbf{ 69.12} & \underline{ 68.94} \\
& 2 shots & 62.61 & 53.72 & 66.53 & 65.72 & 69.03 & 69.92 & 67.1 &69.31     & \underline{ 70.63} & \textbf{ 71.23} \\
& 4 shots & 62.61 & 63.02 & 69.97 & 67.54 & 70.21 & 72.22 & 70.67 &71.72     & \underline{ 73.01} & \textbf{ 73.84} \\
& 8 shots & 62.61 & 69.08 & 71.53 & 69.19 & 70.84 & 73.83 & 73.23 &\textbf{75.56}     & { 75.22} & \underline{ 75.32} \\
& 16 shots & 62.61 & 73.28 & 74.67 & 70.35 & 72.15 & 76.04 & 75.53 &76.12     & \underline{ 76.73} & \textbf{ 77.05} \\ \hline
\multirow{5}{*}{\textbf{Food101}}  & 1 shot                      & 85.85                                                                         & 43.96                                                                         & 84.33                                                                        & 84.21                                                                          & \underline{ 85.65}                                                                        & 82.21                                                                         & 80.51         &  82.29                                                                 & 85.41                                                                                                        & \textbf{ 85.66} \\
                                  & 2 shots                     & \underline{ 85.85}                                                                         & 61.51                                                                         & 84.41                                                                        & 83.61                                                                          & \textbf{ 86.22}                                                                        & 81.01                                                                         & 81.47                &  83.52                                                           & 85.41                                                                                                         & 85.62 \\
                                  & 4 shots                        & 85.85                                                                         & 73.19                                                                         & 84.47                                                                        & 83.81                                                                          & \textbf{ 86.88 }                                                                       & 83.21                                                                         & 81.77                                                                 & 84.09           & 85.81                                                                                                       & \underline{ 85.88} \\
                                  
                                  & 8 shots                       & 85.85                                                                         & 79.79                                                                         & 82.67                                                                        & 84.61                                                                          & \underline{ 86.97}                                                                        & 84.61                                                                         & 83.61                                                                  &  85.34         & 86.71                                                                                                       & \textbf{ 87.88} \\
                                  
                                   & 16 shots                        & 85.85                                                                         & 82.91                                                                         & 84.21                                                                        & 85.91                                                                          & 87.25                                                                        & 86.51                                                                         & 85.33                                                               &  86.43            & \underline{ 87.31}                                                                                                        & \textbf{ 87.41} \\\hline 
\multirow{5}{*}{\textbf{UCF101}}  & 1 shot                      & 67.57                                                                         & 53.66                                                                         & 71.23                                                                        & 66.82                                                                          & 70.31                                                                        & 71.83                                                                         & 71.83      &  72.18                                                                    & \underline{ 72.71}                                                                                                        & \textbf{ 74.61} \\
                                  & 2 shots                     & 67.57                                                                         & 65.78                                                                         & 73.43                                                                        & 69.19                                                                          & 73.51                                                                        & 74.01                                                                         & 74.61            &  75.76                                                               & \underline{ 77.11}                                                                                                       & \textbf{78.21} \\
                                  & 4 shots                        & 67.57                                                                         & 73.28                                                                         & 77.11                                                                         & 73.76                                                                          & 74.82                                                                        & 79.01                                                                         & 78.47          &  79.49                                                                 & \underline{ 81.31}                                                                                                       & \textbf{ 81.66} \\
                                  
                                  & 8 shots                       & 67.57                                                                         & 79.34                                                                         & 80.21                                                                        & 77.91                                                                          & 77.14                                                                        & \underline{ 83.21}                                                                         & 81.37                                                              & 81.09              & 83.01                                                                                                       & \textbf{ 85.01} \\
                                  
                                   & 16 shots                        & 67.57                                                                         & 82.11                                                                         & 82.23                                                                        & 79.97                                                                          & 78.14                                                                        & 84.51                                                                         & \underline{ 85.03}            &   83.77                                                              & 84.71                                                                                                       & \textbf{ 87.15} \\\hline
\multirow{5}{*}{\textbf{OxfordPets}}  
& 1 shot & 89.11 & 44.06 & 90.37 & 88.03 & \underline{91.27} & 89.31 & 89.10 &88.47     & { 91.02} & \textbf{91.91} \\
& 2 shots & 89.11 & 58.37 & 89.80 & 89.09 & \textbf{92.64} & 89.62 & 90.87 &90.21     & { 90.31} & \underline{ 92.42} \\
& 4 shots & 89.11 & 71.17 & 92.57 & 91.05 & 92.81 & 91.51 & 91.90 &91.86     & \underline{ 92.82} & \textbf{ 93.52 }\\
& 8 shots & 89.11 & 78.36 & 91.27 & 90.91 & \textbf{93.45} & 91.91 & 92.57 &92.73     & { 93.02} & \underline{ 93.21} \\
& 16 shots & 89.11 & 85.34 & 91.87 & 92.11 & \underline{93.34} & 92.22 & 92.83 &93.11     & { 92.92} & \textbf{ 93.82} \\ \hline
\multirow{5}{*}{\textbf{EuroSAT}}  
& 1 shot & 48.43 & 49.23 & 54.93 & 62.12 & 55.33 & 38.4 & \textbf{ 71.82} &57.23     & \underline{ 62.43} & 56.23 \\
& 2 shots & 48.43 & 61.98 & 65.17 & 68.42 & 46.74 & 63.02 & \textbf{ 78.32} &67.06     & 70.08 & \underline{ 75.53} \\
& 4 shots & 48.43 & 77.09 & 70.81 & 72.36 & 65.56 & 63.03 & \underline{ 84.51} &  71.83   & 74.52 & \textbf{ 89.98} \\
& 8 shots & 48.43 & 84.43 & 78.07 & 86.58 & 68.21 & \underline{ 87.82} & 87.73 & 77.56    & 77.12 & \textbf{ 90.22} \\
& 16 shots & 48.43 & 87.21 & 84.93 & 90.78 & 73.32 & \underline{ 92.61} & 92.33 & 82.75    & 86.42 & \textbf{ 94.33} \\ \hline
\end{tabular}
\label{2}
\end{table*}

\subsection{Parameter Table}
\vspace{-0.1cm}
Due to the different characteristics of each dataset, there may be different optimal parameter values. Therefore, we provide the optimal hyper-parameters for each dataset in Table~\ref{333}.

\begin{table*}[t]
\begin{center}
\scriptsize
\caption{The parameters table of DCA method on 11 datasets in few-shot learning.}
\renewcommand{\arraystretch}{1.2} 
\setlength{\tabcolsep}{2mm} 
\scalebox{1}{
\begin{tabular}{lccccccccccc}\hline
\textbf{Hyperparameters} & \textbf{DTD} & \textbf{FGVCAircraft} & \textbf{SUN397} & \textbf{Caltech101} & \textbf{OxfordPets} & \textbf{Food101} & \textbf{Flowers102} & \textbf{UCF101} & \textbf{StanfordCars} & \textbf{ImageNet} & \textbf{EuroSAT} \\\hline
$\alpha$               &  5.0     & 5.0             &   100.0      &  10.0          &    1.0        &  1.0       &  10.0          &   5.0     & 10.0  & 10.0   &  100.0     \\
$\beta$                &  0.5     &  0.01            &  0.01       &    0.01        &   0.1         &  0.5      & 0.5           &  0.1     & 0.5   & 0.1   & 2.0         \\
Learning rate          &  20.0     &  2.0    &  20.0    & 0.2     & 0.02         &  20.0       & 0.002           &  20.0      & 0.2   & 2.0  & 20.0         \\ \bottomrule

\end{tabular}}
\label{333}
\end{center}
\end{table*}